\documentclass[runningheads]{llncs}
\usepackage{graphicx}
\usepackage{amsfonts}
\usepackage{amsmath}
\usepackage{color}
\usepackage{multirow}
\usepackage{comment}
\usepackage{booktabs}

\begin{document}

\title{Self-supervised motion descriptor for\\cardiac phase detection in 4D CMR based on\\discrete vector field estimations}





\titlerunning{Cardiac phase detection in 4D CMR}
%
\author{Sven Koehler\inst{1,2}\orcidID{0000-0003-4989-8766} \and
Tarique Hussain\inst{3} \and
Hamza Hussain\inst{1}\and
Daniel Young \inst{3}\and
Samir Sarikouch \inst{4,5,6} \and
Thomas Pickardt \inst{4} \and
Gerald Greil \inst{3} \and
Sandy Engelhardt \inst{1,2}\orcidID{0000-0001-8816-7654}}
\authorrunning{Koehler et al.}

\institute{
Department of Internal Medicine III, Group Artificial Intelligence in Cardiovascular Medicine, Heidelberg University Hospital, D-69120 Heidelberg, Germany \and
DZHK (German Centre for Cardiovascular Research), Heidelberg, Germany \and
Department of Pediatrics, Division of Cardiology; Department of Radiology; Adv. Imaging Research Center, UT Southwestern Medical Center, Dallas, TX, USA \and
German Competence Network for Congenital Heart Defects \and
DZHK (German Centre for Cardiovascular Research), Berlin, Germany \and
Department of Cardiothoracic, Transplantation and Vascular Surgery, Hannover Medical School, Hannover, Germany}
\maketitle              

\begin{abstract}
Cardiac magnetic resonance (CMR) sequences visualise the cardiac function voxel-wise over time. Simultaneously, deep learning-based deformable image registration is able to estimate discrete vector fields which warp one time step of a CMR sequence to the following in a self-supervised manner.
However, despite the rich source of information included in these 3D+t vector fields, a standardised interpretation is challenging and the clinical applications remain limited so far.
In this work, we show how to efficiently use a deformable vector field to describe the underlying dynamic process of a cardiac cycle in form of a derived 1D motion descriptor. 
Additionally, based on the expected cardiovascular physiological properties of a contracting or relaxing ventricle, we define a set of rules that enables the identification of five cardiovascular phases including the end-systole (ES) and end-diastole (ED) without usage of labels.
We evaluate the plausibility of the motion descriptor on two challenging multi-disease, -center, -scanner short-axis CMR datasets. First, by reporting quantitative measures such as the periodic frame difference for the extracted phases. Second, by comparing qualitatively the general pattern when we temporally resample and align the motion descriptors of all instances across both datasets.
The average periodic frame difference for the ED, ES key phases of our approach is $0.80\pm{0.85}$, $0.69\pm{0.79}$ which is slightly better than the inter-observer variability ($1.07\pm{0.86}$, $0.91\pm{1.6}$) and the supervised baseline method ($1.18\pm{1.91}$, $1.21\pm{1.78}$). Code and labels will be made available on our GitHub repository. \url{https://github.com/Cardio-AI/cmr-phase-detection}
\end{abstract}


\section{Introduction}

Analysis of cardiac function is tremendously important for diagnosis and monitoring of cardiac diseases. Therefore, modalities such as CMR are predominantly time-resolved for assessing global pump function and local contractility. 
Identification of key phases in such sequences are beneficial to determine the underlying myocardial mechanics with respect to contraction and relaxation. The end-diastolic (ED) and end-systolic (ES) phases are used to compare standardized clinical measures of the cardiac function, such as the left ventricular (LV) ejection fraction and peak systolic strain. An exact identification of these phases is crucial and has a major impact on the accuracy of strain measurements as shown by Mada et al. \cite{MADA2015148}. 
In addition, Zolgharni et al. \cite{zolgharni2017automatic} recognised a median disagreement of three frames between different observers in manually detecting the ED/ES frames. 
A better phase detection is feasible by evaluating the QRS-complex from electrocardiogram (ECG) signals. However, there is often no ECG-signal permanently stored.

\begin{figure}[t]
  \includegraphics[width=1.\linewidth]{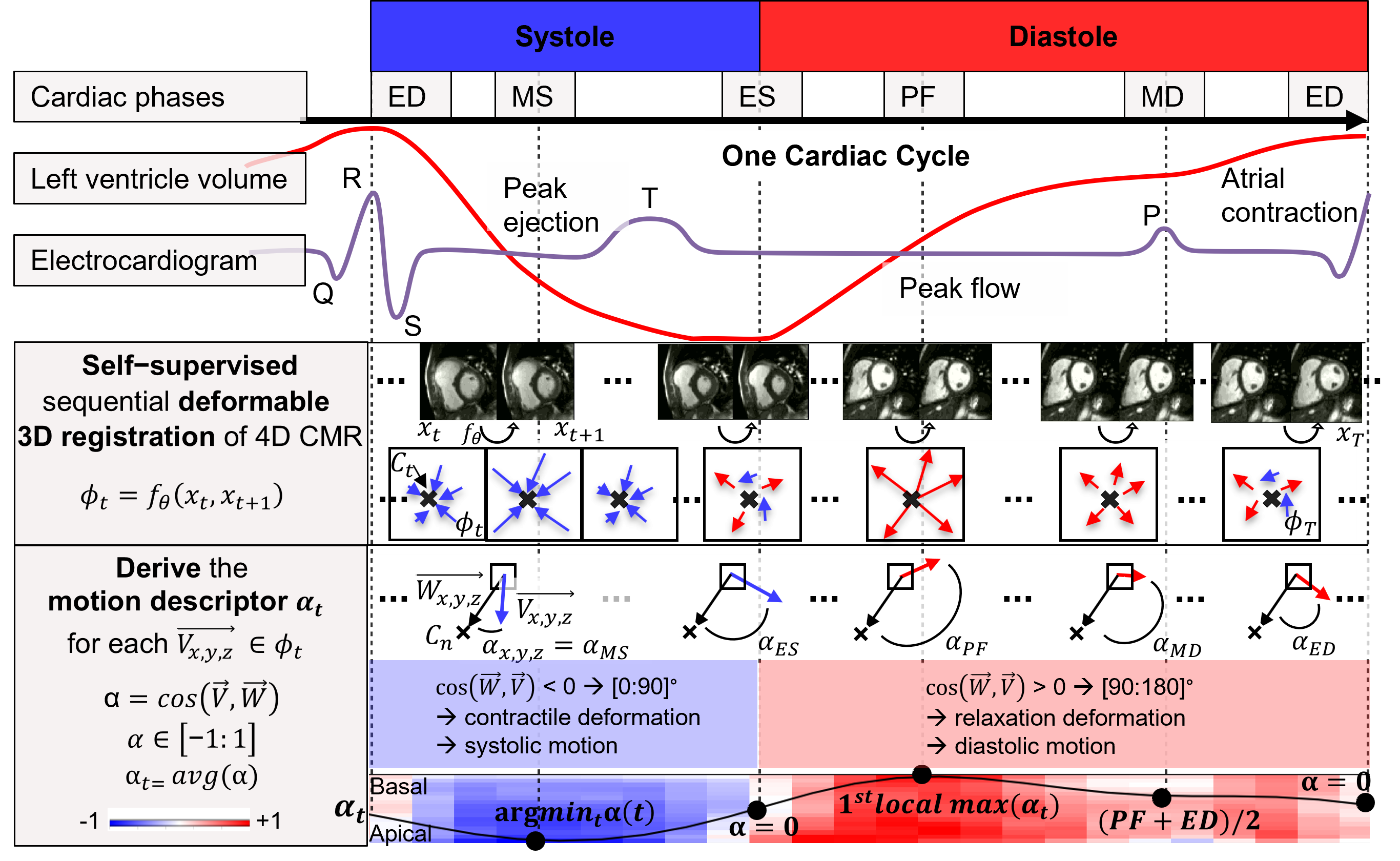}
  \centering
  \caption{Comparison of our self-supervised motion descriptor $\alpha_{t}$ together with a electrocardiogram (ECG) and left ventricular (LV) volume curve over time and the corresponding five labeled key phases ED, MS, ES, PF and MD (cf. Sec. \ref{sec:labels}). First: slice-wise average of $\alpha_t$ - color coded as blue: movement away from the focus point, red: movement towards focus point. Second: volume-wise average as interpolated black curve together with the defined set of rules for each key phase.}
  \label{fig:wiggers_overview}
\end{figure}
Several automatic ECG-free detection approaches have been already suggested. Most of them used echo \cite{darvishi2013measuring,dezaki2018cardiac,Dominguez05classification,fiorito2018detection,gifani2010automatic,kachenoura2007automatic,shalbaf2015echocardiography} or CMR image sequences \cite{kong2016recognizing,Xue2018}.
Early semi-automatic works \cite{darvishi2013measuring,Dominguez05classification,kachenoura2007automatic} required either manual definition of landmarks, the ED frame or an initial contour.

Gifani et al. \cite{gifani2010automatic} and Shalbaf et al. \cite{shalbaf2015echocardiography} solved the problem by using nonlinear dimensionality reduction techniques such as locally linear embedding and isomap.
Evaluation was conducted on small echo cohorts of 8 and 32 patients. 

Recently, fully automatic deep learning based methods for CMR \cite{kong2016recognizing,Xue2018}, echo \cite{dezaki2018cardiac,fiorito2018detection} and fluoroscopy with contrast agent 
\cite{Ciusdel2020} evolved. All of these approaches are supervised, e.g., they require target labels. 
Three out of five predicted the ED and ES 'key frame' directly \cite{Ciusdel2020,dezaki2018cardiac,kong2016recognizing}, in contrast to Xue et al. \cite{Xue2018} and Fiorito et al. \cite{fiorito2018detection}, who classified each frame of the sequence into either diastole or systole. Another obvious approach is to use a LV segmentation approach \cite{Koehler2020} and infer min/max volume from it as ES and ED.

Besides their very promising results, there are limitations. Similar in spirit as a segmentation-based approach, the method of Dezaki and Kong et al. \cite{dezaki2018cardiac,kong2016recognizing} posed the task in a way to detect the maximum/minimum of a regressed LV volume curve. 
However, this method is based on a synthetic volume curve which is only at two points related to the underlying CMR. 
Furthermore, Kong et al. \cite{kong2016recognizing} evaluated the approach on a homogeneous in-house database with CMR sequences of unique lengths that starts with the ED phase and achieved an impressive average frame difference (aFD) of 0.38/0.44 (ED,ES). 
However, their loss definition assumes CMR sequences to start with the ED phase, which is not the case in our multi-centric datasets.
The methodological extension by Dezaki et al. \cite{dezaki2018cardiac} was evaluated on a bigger cohort of echo sequences with varying length, collected from one clinic. Their supervised approach achieved an $aFD$ of 0.71/1.92 (ED/ES), with the mitral valve as strong indicator for the ES phase, visible in their modality. Fiorito et al. \cite{fiorito2018detection} defined a binary classification task to detected two phases on echo sequences and achieved an $aFD$ of 1.52, 1.48 (ED/ES).
All fore-mentioned deep-learning approaches are supervised and need either labels of the left ventricle (LV)-blood pool \cite{dezaki2018cardiac,kong2016recognizing} or cardiac phase labels \cite{fiorito2018detection}. In addition, they rely on the LV-volume change over time for cardiac phase estimation, which is in some aspects an assumption that does not always hold true:  For example, in the isovolumetric contraction and relaxation phases around ED and ES the myocardium changes without affecting ventricular volumes. Therefore, our hypothesis is that phase detection based on myocardial deformation might be more accurate than using LV volume change as a marker.

In this work, we show how to efficiently reduce a 3D+t deformable vector field into a 1D motion descriptor representing the general dynamic pattern of a cardiac cycle in a self-supervised manner (cf. Fig.\ref{fig:wiggers_overview}). We evaluate the plausibility of this motion descriptor and apply it to the task of cardiac phase detection. Here, we define a rule-set based on the expected cardiovascular physiological properties of a contracting/relaxing ventricle and extract the ED/ES phase at the turning point of deformation. To verify the plausibility of the motion descriptor in the intermediate frames, we further extend this rule-set to extract three additional physiologically relevant time points (cf. Sec. \ref{eq:ruleset}) and compare them with clinical labels from an experienced pediatric cardiologist on two heterogeneous cohorts.

\section{Methods}

This work is based on the assumption that a sequential deformable volume-to-volume registration field roughly represent the dynamics of the heart along a cardiac cycle. However, such 3D+t vector fields might become large, they are often spatially not aligned between patients and represent also deformation of the surrounding area which makes an automatic interpretation in a patient cohort challenging. Therefore, our rationale is to compress essential information into a 1D motion descriptor, which is decorrelated from the image grid.
\subsection{Model definition}

Let a 3D+t CMR sequence be defined as $x$ where each cardiac volume $x_t$ is a 3D image at time point $t=\{1,...,T\}$.
We defined a deformable registration task (cf. Fig. 1a. in Suppl. Material) 
as $\phi, \hat{M} = f_{\theta}(M,F)$ with $M$, $F$ the moving and fixed volume pair, $\theta$ the learnable parameters of $f$, $\phi$ the resulting discrete vector field and $\hat{M}$ the moved volume after $\phi$ was applied with a spatial transformer layer to $M$. 
In our 4D case, for $M={x}_{t}$ and $F=x_{t+1}$ we obtain $\phi_t, \hat{x}_{t+1} = f_{\theta}(x_{t},x_{t+1})$. Additionally, due to the periodic behaviour, the last volume of the sequence $x_T$ is warped to the first $x_1$. Therefore, $\phi_T$ is defined as $\phi_T,\hat{x}_{1} = f_{\theta}(x_{T},x_{1})$.

We define one 3D focus point $C_n \in \mathbb{Z}^{3}$ that we use for each volume $x_t$ (cf. Fig. 1b. in Suppl Material), which can represent an anatomical landmark or a point that is calculated without prior knowledge (cf. Sec. \ref{sec:experiments}). For each vector $\vec{v} \in \phi_t$, we calculate the angle $\alpha = cos(\vec{v}, \vec{w})$, where the vector $\vec{w}$ points from the corresponding $x,y,z$ grid position to $C_n$. Therefore, $\vec{v}$ with $\alpha \in [-1,0[$ will point towards the focus point $C_n$ and with $\alpha \in ]0,1]$ away from it. This enables voxel-wise differentiation of a contractile deformation vector from a vector that describes relaxation, which is the main rationale we followed for phase extraction.

Additionally, we calculate the 70th quantile on the temporally averaged euclidean norm $|\vec{v}|$ of $\phi$ and use it as threshold to filter non-cardiac motion information and noise from the cardiac deformation (cf. row d. in Fig. \ref{fig:examples}).

Finally, for the extraction of the 1D curves over time (motion descriptor $\alpha_t$ and $|\vec{v}|_t$) we average $\alpha$ and $|\vec{v}|$ per 3D volume, 
apply a Gaussian filter ($\sigma = 2$) on $\alpha_t$ and min/max normalise both into a range of $[-1,1]$ and $[0,1]$, respectively. The scaling of $\alpha_t$ may introduce small shifts ($<$1 frames) to the zero-crossing time points $t_{ED}$ and $t_{ES}$ but also removes diffuse zero crossings for weak pathological relaxation phases.

\subsection{Loss function and key frame extraction}
The registration loss consists of an image similarity component $\mathcal{L}_{sim}$ and a regularizer $\mathcal{L}_{smooth}$ and is defined by Eq. \ref{eq:loss}. It turned out that the structural similarity index measure (SSIM) \cite{wang_ssim_2004}, which is based on a luminance, contrast and structure measurement, performs better than the mean squared error as $\mathcal{L}_{sim}$. Here, we average the 2D SSIM per 3D volume, the equal weighted general 2D form is given by Eq. \ref{eq:ssim} with $\mu_x$, $\mu_y$ as the average and $\sigma_x^2$, $\sigma_y^2$ as the variance of a $N \times N$ region in $x$ and $y$, which correspond to the two neighbouring time steps. $\sigma_{xy}$ is given by the co-variance of $x$ and $y$ and $\epsilon_1$, $\epsilon_2$ are two variables included to avoid instability. For $\mathcal{L}_{smooth}$ (cf. Eq. \ref{eq:smooth}) of $\phi$ we used the same diffusion regularizer of the spatial gradients as introduced by Balakrishnan et al.\cite{balakrishnan2018reg}, with $\Omega$ representing each voxel in $x_t$ and set the regularization parameter $\lambda= 0.001$.

\begin{align}
\mathcal{L}(F,M,\phi) &=\mathcal{L}_{sim}(F,M(\phi)) + \lambda \mathcal{L}_{smooth}(\phi).\label{eq:loss}\\
{SSIM}(x,y) &= \frac{(2\mu_x\mu_y + \epsilon_1)(2\sigma_{xy} + \epsilon_2)}{(\mu_x^2 + \mu_y^2 + \epsilon_1)(\sigma_x^2 + \sigma_y^2 + \epsilon_2)}.
    \label{eq:ssim}\\
\mathcal{L}_{smooth}(\phi) &= \sum_{p\in \Omega}||\nabla \phi(p)||^2.
\label{eq:smooth}
\end{align}

\label{sec:labels}
We define a set of rules that derives five cardiac key time points from the compressed 1D signal $\alpha_t$, including the ED, mid-systole (MS; maximum contraction resulting in a peak ejection between ED and ES), ES, peak flow (PF; peak early diastolic relaxation) and mid-diastole (MD; phase before atrial contraction at the on-set of the p-wave). As shown in Fig. \ref{fig:wiggers_overview}, the succession of these labels is temporally correlated, however, we observed that the starting cardiac phase of the CMR sequences varies between acquisitions (cf. Sec. \ref{sec:data}). In order to handle CMR sequences with varying starting phases we first detected the time-point with maximum contraction (MS) and applied the other rules sequentially to the cyclic sub-sequences. Please note that this rule-set (especially for the MD rule, where we could also rely on the last peak as indicator for the atrial contraction) is a work-in-progress-trade-off between accuracy and the generalisation capability towards cut-off or pathological sequences. For this work we preferred simple rules that are based on a physiological reasoning over complicated ones. Especially, the intermittent diastolic pattern of $\alpha_t$ differs between pathological patients and sometimes leads to multiple diffuse relaxation peaks, which are difficult to assign to the presumed peak flow or to the atrial contraction close to the MD phase.

\begin{align}
    MS &= t_{m} \text{ with } \alpha(t_{m}) \leq \alpha(t) &\text{with } t\in T\nonumber\\
    ES &=  \min_t \alpha(t) = 0 &\text{with } t\in [MS;PF]\nonumber\\
    PF &=  \min_t \alpha^{\prime}(t)=0 \wedge \alpha^{\prime\prime}(t)<0 &\text{with } t\in [ES;MS]\label{eq:ruleset}\\
    ED &=  \max_t \alpha(t) = 0 &\text{with } t\in [PF;MS] \nonumber\\
    MD &= (PF + ED)/2 &\nonumber
\end{align}

\subsection{Deep Learning Framework}
Our Deep Learning model consists of a 3D CNN-based deformable registration-module followed by the direction-module.
For the sequential volume-to-volume deformable registration we use a slightly modified time distributed 3D U-Net as introduced by Ronneberger et al. \cite{Ronneberger2015} followed by a spatial transformer layer, such as used by Balakrishnan et al. \cite{balakrishnan2018reg}. For further details we refer to our GitHub repository.
Our final model expects a 4D volume as input with $b\times 40 \times 16 \times64 \times64$ as batchsize, time, spatial slices and x-/y dimension. Subsequently, the direction-module calculates voxel-wise $\alpha$, $|\vec{v}|$ and per 4D volume the 1D curves $\alpha_t$ and $|\vec{v}|_t$ which are used in our rule-set. Please note: all parts are differentiable, learning is end-to-end and could be used in a supervised approach.

\subsection{Datasets}
\label{sec:data}
For evaluation we used two 4D cine-SSFPs CMR SAX datasets. The annotations were made by an experienced paediatric cardiologist. Distributions and mean occurrence of the phases are shown in Tab 1 and Fig.3a. in the Suppl. Material.

First, the publicly available \textit{Automatic Cardiac Diagnosis Challenge} (ACDC) dataset \cite{Bernard2018a} (100 patients, 5 pathologies, 2 centers) was used. 
The mean$\pm{SD}$ number of frames is $26.98\pm{6.08}$, within a range of $[12, 35]$. Furthermore, not all 4D sequences capture an entire cardiac cycle (cf. Fig. 2 in Suppl Material).
A detailed description of the dataset can be found in \cite{Bernard2018a} and in Sec. 1 of our Suppl. Material. We corrected and extended the original cardiac phase labels (e.g.: the ED phase was uniformly labelled at frame 0 over the entire cohort, which is a rough approximation). Now, 75 sequences start close to the MS and 25 close to the ED phase. Our labels will be released on our GitHub repository. The inter-observer error between the original phase labels and our re-labelling is $0.99\pm{1.23}$, with a maximal distance of $6$ (ED) and $10$ (ES), respectively.

Second, a multi-centric dataset (study identifier: NCT00266188) \cite{KoehlerTMI2021,Koehler2020,Sarikouch2011} of 278 patients with Tetralogy of Fallot (TOF), which is a complex congenital heart disease, was used. 
The mean number of frames is $21.92\pm{4.02}$, within a range of $[12, 36]$. The sequence length of each cardiac cycle is 743$\pm{152}$ms, within a range of $[370,1200]$. 
191 sequences start close to the MS and 84 close to the ED phase. The other three phases occurred once at the sequence start.

\section{Experiments}
\label{sec:experiments}
\begin{figure}[t]
  \includegraphics[width=1.\linewidth]{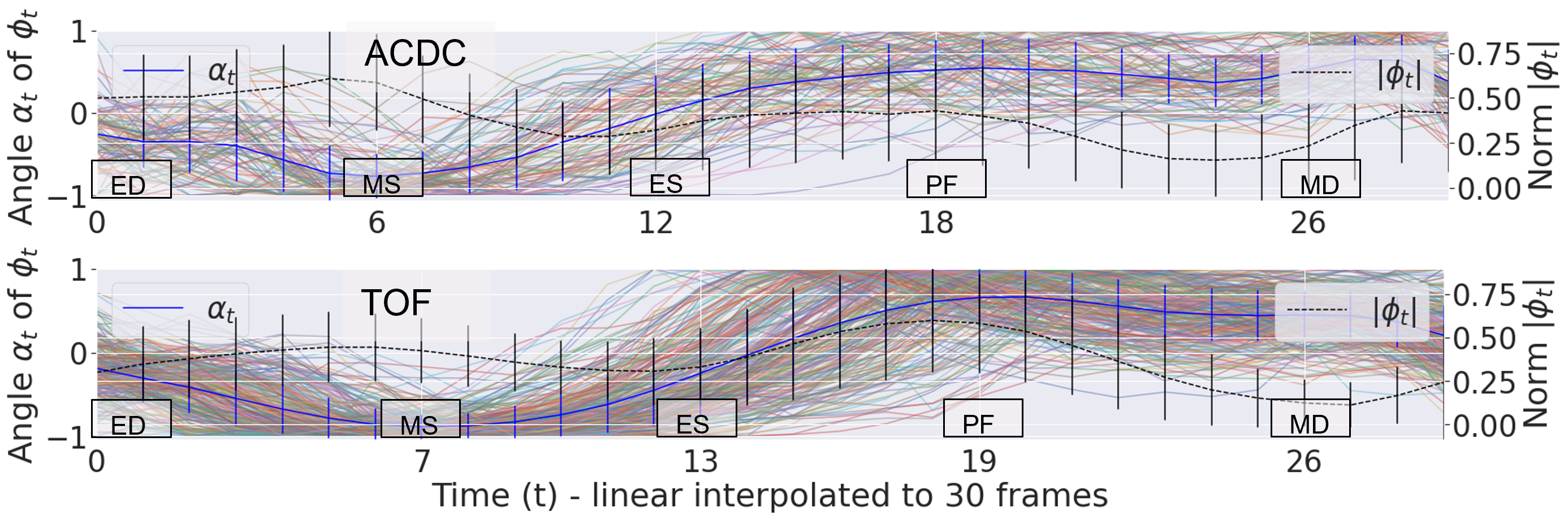}
  \centering  \caption{Qualitative results for the motion descriptor $\alpha_t$ on both datasets. Both plots show the per-cohort average of $\alpha_t$ (blue/left axis) and $|\vec{v}|_t$ (black/right axis) together with the temporal aligned, resampled and averaged phase indices (x-axis), without any post-processing. We overlaid the per-instance curves of $\alpha_t$ but for the sake of clarity, we omitted the instances curves for $|\vec{v}|_t$. Please note: $\alpha_t<0$ correspond to systolic and $\alpha_t >0$ to diastolic frames. The peaks in $|\vec{v}|_t$ are close to the mid-systolic (MS) and peak flow (PF) phases. We aligned/resized the data to visualise the general properties of $\alpha_t$ and $|\vec{v}|_t$, while the original data for training or inference remained unaligned.
  } 
  \label{fig:aligned_dir_norm}
\end{figure}

We extend the previously used \cite{dezaki2018cardiac,fiorito2018detection,gifani2010automatic,kong2016recognizing,shalbaf2015echocardiography} average Frame Difference ($aFD = |p_i - \hat{p_i}| \label{eq:aFD}$)  to account for the periodicity of the cardiac cycle, 
and refer to it as 
\begin{align}
    pFD(p_i,\hat{p_i}) &= \min(|p_i - \hat{p_i}|, T - \max(p_i,\hat{p_i}) + \min(p_i,\hat{p_i})) \label{eq:pFD}
\end{align}
with $i \in [ED,MS,ES,PF,MD]$ and $p_i$, $\hat{p_i}$ the $i$th ground truth and predicted label. This is important for permuted sequences when the annotated phase $p_i$ is labelled at $t=1$ but, $\hat{p_i}$ predicts $t=T$ and vice versa. The $pFD$ would be 1; in the original $aFD$ formulation, the distance would be $T$.

Each experiment was carried out in a four-fold-cross validation manner.
We resampled $x$ with linear interpolation to a spacing of $2.5$ mm$^{3}$ and repeated $x_t$ along $t$ until we reached the network's input size of 40. Following that, we focus crop with different focus points $C_n$ in 3D (cf. next paragraph), clipped outliers with a quantile of $.999$ and standardised per 4D. We did not apply any image-based augmentation as we noticed no over-fitting in $\mathcal{L}_{sim}$.
We compare our results with a supervised LV-volume-based approach on the same data and refer to it as $base$. Four U-net based segmentation models were trained on the public ACDC data (LV DICE: 0.91 ± 0.02). Next, we applied a connected component filter and identified the ED/ES frames based on the min/max LV volume. Later we applied one of them on the TOF dataset to provide a supervised baseline.

\begin{figure}[t]
  \includegraphics[width=1\linewidth]{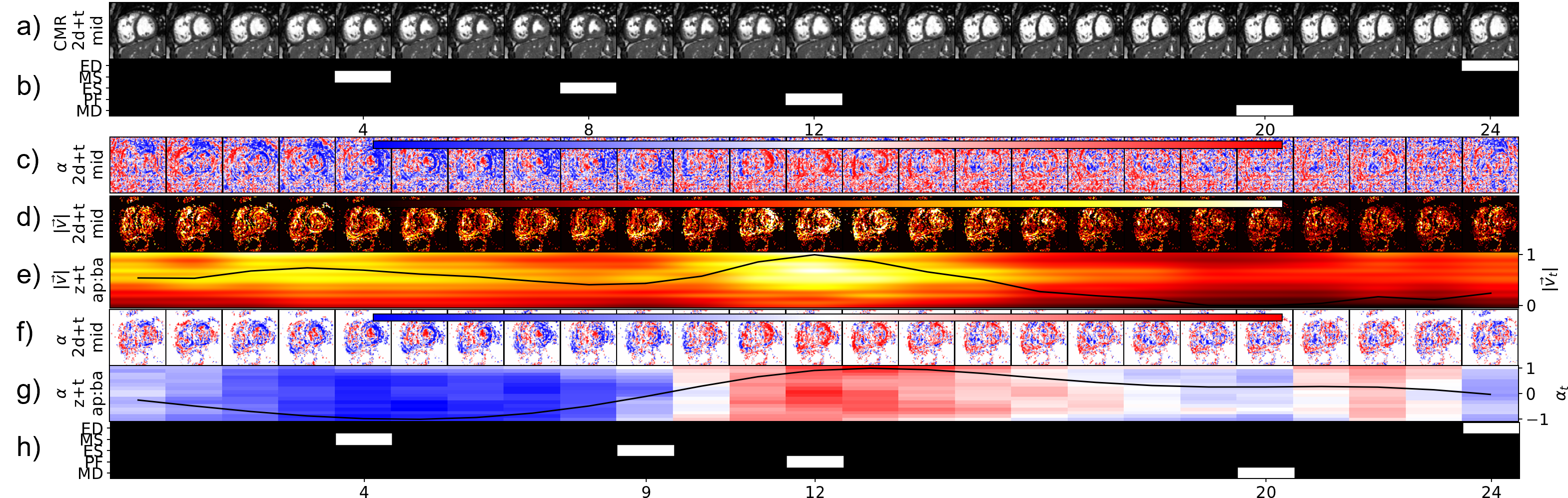}
  \centering
  \caption{Visualisation for one cardiac cycle of a random TOF patient. 
  a) Mid-cavity CMR slice, 
  b) GT phase labels, 
  c) Mid-cavity view of the direction $\alpha$ without percentile masking (blue: towards focus point, red: away from),
  d) Mid-cavity view of $|\vec{v}|$ (masked by the 70th percentile of $|\vec{v}|$),
  e) Color coded slice-wise average of $|\vec{v}|$ (base2apex) in background. Volume-based average of $|\vec{v}|_t$ as curve with axis on the right,
  f) Mid-cavity view of $\alpha$ (same color-coding, masked by the 70th percentile of $|\vec{v}|$),
  g) Color coded slice-wise average of $\alpha$ (base2apex) in background. Volume-based average of $\alpha_t$ as curve with axis on the right,
  h) Predicted phase.}
  \label{fig:examples}
\end{figure}

Quantitatively, we report the $pFD$ (cf. Eq. \ref{eq:pFD}) per dataset and cardiac key-point in the original temporal resolution in Table \ref{tab:results_afd}.
Additionally, we investigate the sensitivity of different $C_n$ on the $pFD$ and compare the LV blood-pool center of mass $C_{lv}$, the mean septum landmark (center between the average anterior and inferior right ventricular insertion points (RVIP) \cite{KoehlerBVM2022}) $C_{sept}$, the CMR-volume center $C_{vol}$ and the center of mass for a quantile-threshold mean squared error mask averaged along the temporal axis $C_{mse}$.
Finally, we qualitatively evaluate the general pattern of $\alpha_t$ and $|\vec{v}|_t$ on both datasets (cf. Fig. \ref{fig:aligned_dir_norm}) and visualise different views for one random patient (cf. Fig. \ref{fig:examples}).

\section{Results}

Based on the per-instance and global average curves in Fig. \ref{fig:aligned_dir_norm}, we provide a qualitative and more clinical interpretation of our results. The general systolic/diastolic pattern of $\alpha_t$ across all patients of both cohorts aligns with our physiological expectations and show a negative course (mainly contractile direction of $\phi$) during the systole followed by multiple positive local maxima during the diastole. Furthermore, we would expect a change from contractile to relaxile deformation (ES) or vice versa (ED) where $\alpha_t$ has a zero pass. From our observations and confirmed by the $pFD$ in Tab. \ref{tab:results_afd}, this is usually true.

In general, $\alpha_t$ shows a clear global minimum close to the MS-phase, that refers to the time-point where most of the masked voxels (percentile-based threshold of $|\vec{v}|$) represent the greatest contractile deformation direction. The relaxation part is often more diffuse but with mostly two peaks in $\alpha_t$ (the systolic negative course is greater than the diastolic positive course if we omit the re-scaling).
The time point of minimal and maximal deformation should correspond to the peaks and valleys as shown in Fig. \ref{fig:aligned_dir_norm} for the cohort-based average of $|\vec{v}|_t$. On both datasets we have a local maxima of $|\vec{v}|_t$ close to the MS during systole and close to the PF or shortly after the MD phases during diastole. The global minimum of $|\vec{v}|_t$ is often around the MD phase (cf. Fig. \ref{fig:aligned_dir_norm} and Fig. \ref{fig:examples}). 
From a visual point of view, the threshold $|\vec{v}|_t$ mask is able to eliminate most of the non-cardiac information (cf. row f. in Fig. \ref{fig:examples}).

As quantitative measure we present the $pFD$ (cf. Eq. \ref{eq:pFD}) for both cohorts and for each phase
in Table \ref{tab:results_afd}. In addition we report the error distribution per key time point in Fig. 3b in our Suppl. Material. The cardiac contraction happens usually in one coherent contractile deformation, which results in a clear negative course of $\alpha_t$, that makes phase extraction straight-forward, which is visible in lower $pFD$ scores for the key-points (ED, MS, ES). This is in contrast to the relaxation of the heart, which does not follow such an homogeneous pattern. In fact, we observed multiple peaks that may result from basal to apical regions relaxing at different rate (cf. Fig. \ref{fig:aligned_dir_norm}). The $pFD$ for the diastolic phases (PF and MD) are slightly worse and represent the difficulties to assign these peaks to either the ventricle contraction during the peak flow (PF) or to the atrial contraction shortly after the mid-diastolic (MD) phase. Both experiments that are based on prior knowledge ($C_{lv}$ and $C_{sept}$) provided similar results except for the PF and MD phases, where both performed once better. Our experiment $C_{vol}$, results, as expected, in the highest $pFD$ and $SD$. Using the center of mass of the temporally sequential mean squared error of $x_t$ and $x_{t+1}$ as focus point $C$ (cf. Sec. \ref{sec:experiments}) closed the gap of the unsupervised approaches with similar or slightly better $pFD$ while removing the need of prior knowledge or labels.

\begin{table}
\centering
\caption{Sensitivity of the resulting $pFD$ for two cohorts with respect to different focus points $C_n$ (cf. Sec. \ref{sec:experiments}) and comparison to supervised segmentation-based approach $base$.\\
$^1$ = $C$ based on anatomical GT knowledge, \\ $^2$ = $C$ based on more generic information (unsupervised).
}
\begin{tabular*}{\textwidth}{lccccccc}
Data &$C_n$         &all          &ED             &MS                 &ES             &PF             &MD\\
\toprule
ACDC &$base$      & -  &$1.13\pm{1.82}$ & -  &$\textbf{0.95}\pm{1.29}$ & -  & - \\
   &$C_{lv}^1$      &$1.36\pm{1.37}$ &$1.13\pm{1.23}$ &$\textbf{0.97}\pm{0.95}$ &$1.05\pm{1.09}$ &$1.87\pm{1.98}$ &$1.77\pm{1.59}$\\
   &$C_{sept}^1$    &$1.32\pm{1.21}$ &$1.09\pm{1.09}$ &$\textbf{0.97}\pm{0.83}$ &$0.96\pm{0.87}$ &$1.68\pm{1.63}$ &$1.91\pm{1.65}$\\
   &$C_{vol}^2$     &$1.56\pm{1.86}$ &$1.37\pm{2.01}$ &$1.24\pm{1.40}$ &$1.19\pm{1.60}$ &$1.99\pm{2.14}$ &$2.01\pm{2.15}$\\
   &$C_{mse}^2$     &$\textbf{1.29}\pm{1.25}$ &$\textbf{1.08}\pm{1.26}$ &$1.02\pm{0.94}$ &$0.97\pm{0.95}$ &$\textbf{1.66}\pm{1.56}$ &$\textbf{1.73}\pm{1.54}$\\
\midrule
TOF  &$base$      & - &$1.18\pm{1.91}$ & - &$1.21\pm{1.78}$ & - & - \\
    &$C_{lv}^1$       &$0.99\pm{0.91}$ &$0.81\pm{0.93}$ &$1.07\pm{0.79}$ &$0.72\pm{0.79}$ &$0.90\pm{0.82}$ &$\textbf{1.46}\pm{1.22}$\\
    &$C_{sept}^1$     &$\textbf{0.95}\pm{0.89}$ &$0.82\pm{0.88}$ &$\textbf{0.87}\pm{0.72}$ &$0.70\pm{0.76}$ &$\textbf{0.78}\pm{0.83}$ &$1.58\pm{1.26}$\\
    &$C_{vol}^2$      &$1.02\pm{0.97}$ &$0.86\pm{1.04}$ &$1.06\pm{0.83}$ &$0.76\pm{0.80}$ &$0.88\pm{0.90}$ &$1.56\pm{1.28}$\\
    &$C_{mse}^2$      &$0.97\pm{0.91}$ &$\textbf{0.80}\pm{0.85}$ &$0.94\pm{0.76}$ &$\textbf{0.69}\pm{0.79}$ &$0.85\pm{0.86}$ &$1.57\pm{1.27}$\\
\bottomrule
\end{tabular*}
\label{tab:results_afd}
\end{table}

\section{Discussion and Conclusion}

In this work we compute a motion descriptor based on the mean direction and norm of a sequential deformable registration field $\phi_t$ in a self-supervised manner according to different focus points $C_n$, to derive the cardiac dynamics over time.

Furthermore, according to the expected properties of a vector field that mainly represents myocardial contraction and relaxation, we define a set of rules and extend the state-of-the-art by extracting not only two but five cardiovascular key-time frames on CMR sequences with any length and independent of the starting phase. To the best of our knowledge this has not been done before.
We evaluate the reliability of the motion descriptor on two challenging multi-center datasets and compare our method to a supervised baseline. Even though the set of rules was defined empirically, we could quantitatively and qualitatively confirm that the self-supervised motion descriptor $\alpha_t$ is able to express the expected, underlying cardiovascular physiological motion properties. We will release our extended ACDC phase labels to enable future comparison.

The $pFD$ (ED,ES) of the completely self-supervised experiment (TOF, $C_{mse}$: 0.80$\pm{0.85}$, 0.69$\pm{0.79}$) is slightly better than the recognised inter-observer error (ACDC: 1.07$\pm{0.86}$, 0.91$\pm{1.6}$) and significantly ($p<0.001$) 
better than the supervised baseline (TOF, $base$: 1.18$\pm{1.91}$, 1.21$\pm{1.78}$).

Supervised methods may achieve promising/comparable results (cf. 1st row ACDC in Table \ref{tab:results_afd}: 1.13$\pm{1.82}$,0.95$\pm{1.29}$), nevertheless their performance often drops (ES: -$0.26$) when they are applied to unseen datasets, due to the inherent domain shift (cf. 2nd row TOF in Table \ref{tab:results_afd}). This is were self-supervised methods might unfold their strengths, since model re-training or adjustment does not rely on annotations and can be easily done on domain shifted data.

This work assumes that CMR sequences capture an entire cardiac cycle. Cut-off sequences may result in unphysiological peaks in $|\vec{v}|_t$ and hence slightly worse results for phase detection (cf. ACDC dataset in Table \ref{tab:results_afd}). However, we show how to benefit from these information to automatically detect cut-off sequences in a self-supervised way for quality control purposes (cf. Suppl. Material Fig. 2). These cut-off sequences refrained us from using the magnitude directly as key-frame detector, e.g. identifying the peak ejection/peak flow phases based on the largest overall vector magnitudes. In one experiment we weight $\alpha$ of each voxel by the corresponding magnitude $|\vec{v}|$, unfortunately this descriptor performed worse for cardiac phase detection. 
In future work, we will show the value of this descriptor for inter-patient comparison and cardiovascular pathology description.

\subsubsection{Acknowledgments}
This work was supported in parts by the Informatics for Life Project through the Klaus Tschira Foundation, 
by the Competence Network for Congenital Heart Defects (Federal Ministry of Education and Research/grant number 01GI0601) and the National Register for Congenital Heart Defects (Federal Ministry of Education and Research/grant number 01KX2140),
by the German Centre for Cardiovascular Research (DZHK) 
and the SDS@hd service by the MWK Baden-Württemberg and the DFG through grant INST 35/1314-1 FUGG and INST 35/1503-1 FUGG.

\bibliographystyle{splncs04}

\bibliography{references_new.bib}

\begin{thebibliography}{10}
\providecommand{\url}[1]{\texttt{#1}}
\providecommand{\urlprefix}{URL }
\providecommand{\doi}[1]{https://doi.org/#1}

\bibitem{balakrishnan2018reg}
Balakrishnan, G., Zhao, A., Sabuncu, M., Guttag, J., Dalca, A.V.: An
  unsupervised learning model for deformable medical image registration. CVPR:
  Computer Vision and Pattern Recognition pp. 9252--9260 (2018)

\bibitem{Bernard2018a}
Bernard, O., Lalande, A., Zotti, C., Cervenansky, F., Yang, X., Heng, P.,
  et~al.: Deep {Learning} {Techniques} for {Automatic} {MRI} {Cardiac}
  {Multi}-{Structures} {Segmentation} and {Diagnosis}: {Is} the {Problem}
  {Solved}? IEEE Transactions on Medical Imaging  \textbf{37}(11),  2514--2525
  (2018). \doi{10.1109/TMI.2018.2837502}

\bibitem{Ciusdel2020}
Ciusdel, C., Turcea, A., Puiu, A., Itu, L., Calmac, L., Weiss, E., Margineanu,
  C., Badila, E., Berger, M., Redel, T., et~al.: Deep neural networks for
  ecg-free cardiac phase and end-diastolic frame detection on coronary
  angiographies. Computerized Medical Imaging and Graphics  \textbf{84},
  101749 (2020)

\bibitem{darvishi2013measuring}
Darvishi, S., Behnam, H., Pouladian, M., Samiei, N.: Measuring left ventricular
  volumes in two-dimensional echocardiography image sequence using level-set
  method for automatic detection of end-diastole and end-systole frames.
  Research in Cardiovascular Medicine  \textbf{2}(1), ~39 (2013)

\bibitem{dezaki2018cardiac}
Dezaki, F.T., Liao, Z., Luong, C., Girgis, H., Dhungel, N., Abdi, A.H.,
  Behnami, D., Gin, K., Rohling, R., Abolmaesumi, P., et~al.: Cardiac phase
  detection in echocardiograms with densely gated recurrent neural networks and
  global extrema loss. IEEE transactions on medical imaging  \textbf{38}(8),
  1821--1832 (2018)

\bibitem{Dominguez05classification}
Dominguez, C.R., Kachenoura, N., Mul{\'{e}}, S., Tenenhaus, A., Delouche, A.,
  Nardi, O., G{\'{e}}rard, O., Diebold, B., Herment, A., Frouin, F.:
  Classification of segmental wall motion in echocardiography using quantified
  parametric images. In: Frangi, A.F., Radeva, P., Santos, A., Hernandez, M.
  (eds.) Functional Imaging and Modeling of the Heart, Third International
  Workshop, {FIMH} 2005, Barcelona, Spain, June 2-4, 2005, Proceedings. Lecture
  Notes in Computer Science, vol.~3504, pp. 477--486. Springer (2005).
  \doi{10.1007/11494621\_47}

\bibitem{fiorito2018detection}
Fiorito, A.M., {\O}stvik, A., Smistad, E., Leclerc, S., Bernard, O.,
  Lovstakken, L.: Detection of cardiac events in echocardiography using 3d
  convolutional recurrent neural networks. In: 2018 IEEE International
  Ultrasonics Symposium (IUS). pp.~1--4. IEEE (2018)

\bibitem{gifani2010automatic}
Gifani, P., Behnam, H., Shalbaf, A., Sani, Z.A.: Automatic detection of
  end-diastole and end-systole from echocardiography images using manifold
  learning. Physiological measurement  \textbf{31}(9), ~1091 (2010)

\bibitem{kachenoura2007automatic}
Kachenoura, N., Delouche, A., Herment, A., Frouin, F., Diebold, B.: Automatic
  detection of end systole within a sequence of left ventricular
  echocardiographic images using autocorrelation and mitral valve motion
  detection. In: 2007 29th Annual International Conference of the IEEE
  Engineering in Medicine and Biology Society. pp. 4504--4507. IEEE (2007)

\bibitem{KoehlerTMI2021}
Koehler, S., Hussain, T., Blair, Z., Huffaker, T., Ritzmann, F., Tandon, A.,
  Pickardt, T., Sarikouch, S., Latus, H., Greil, G., Wolf, I., Engelhardt, S.:
  Unsupervised domain adaptation from axial to short-axis multi-slice cardiac
  {MR} images by incorporating pretrained task networks. {IEEE} Transactions on
  Medical Imaging  \textbf{40}(10),  2939--2953 (2021).
  \doi{10.1109/tmi.2021.3052972}

\bibitem{KoehlerBVM2022}
Koehler, S., Sharan, L., Kuhm, J., Ghanaat, A., Gordejeva, J., Simon, N.K.,
  Grell, N.M., Andr{\'e}, F., Engelhardt, S.: Comparison of evaluation metrics
  for landmark detection in cmr images. In: Maier-Hein, K., Deserno, T.M.,
  Handels, H., Maier, A., Palm, C., Tolxdorff, T. (eds.) Bildverarbeitung
  f{\"u}r die Medizin 2022. pp. 198--203. Springer Fachmedien Wiesbaden,
  Wiesbaden (2022)

\bibitem{Koehler2020}
Koehler, S., Tandon, A., Hussain, T., Latus, H., Pickardt, T., Sarikouch, S.,
  et~al.: {How well do U-Net-based segmentation trained on adult cardiac
  magnetic resonance imaging data generalize to rare congenital heart diseases
  for surgical planning?} In: Medical Imaging 2020: Image-Guided Procedures,
  Robotic Interventions, and Modeling. vol. 11315, pp. 409 -- 421.
  International Society for Optics and Photonics, SPIE (2020).
  \doi{10.1117/12.2550651}

\bibitem{kong2016recognizing}
Kong, B., Zhan, Y., Shin, M., Denny, T., Zhang, S.: Recognizing end-diastole
  and end-systole frames via deep temporal regression network. In:
  International conference on medical image computing and computer-assisted
  intervention. pp. 264--272. Springer (2016)

\bibitem{MADA2015148}
Mada, R.O., Lysyansky, P., Daraban, A.M., Duchenne, J., Voigt, J.U.: How to
  define end-diastole and end-systole?: Impact of timing on strain
  measurements. JACC: Cardiovascular Imaging  \textbf{8}(2),  148--157 (2015).
  \doi{https://doi.org/10.1016/j.jcmg.2014.10.010}

\bibitem{Ronneberger2015}
Ronneberger, O., Fischer, P., Brox, T.: {U-Net: Convolutional Networks for
  Biomedical Image Segmentation}. In: Medical Image Computing and
  Computer-Assisted Intervention -- MICCAI 2015. pp. 234--241. Springer
  International Publishing (2015)

\bibitem{Sarikouch2011}
Sarikouch, S., Koerperich, H., Dubowy, K.O., Boethig, D., Boettler, P., Mir,
  T.S., et~al.: {Impact of Gender and Age on Cardiovascular Function Late After
  Repair of Tetralogy of Fallot}. Circulation: Cardiovascular Imaging
  \textbf{4}(6),  703--711 (2011). \doi{10.1161/CIRCIMAGING.111.963637}

\bibitem{shalbaf2015echocardiography}
Shalbaf, A., AlizadehSani, Z., Behnam, H.: Echocardiography without
  electrocardiogram using nonlinear dimensionality reduction methods. Journal
  of Medical Ultrasonics  \textbf{42}(2),  137--149 (2015)

\bibitem{wang_ssim_2004}
Wang, Z., Bovik, A., Sheikh, H., Simoncelli, E.: Image quality assessment: from
  error visibility to structural similarity. IEEE Transactions on Image
  Processing  \textbf{13}(4),  600--612 (2004). \doi{10.1109/TIP.2003.819861}

\bibitem{Xue2018}
Xue, W., Brahm, G., Pandey, S., Leung, S., Li, S.: Full left ventricle
  quantification via deep multitask relationships learning. Medical Image
  Analysis  \textbf{43},  54--65 (2018).
  \doi{https://doi.org/10.1016/j.media.2017.09.005}

\bibitem{zolgharni2017automatic}
Zolgharni, M., Negoita, M., Dhutia, N.M., Mielewczik, M., Manoharan, K.,
  Sohaib, S.A., Finegold, J.A., Sacchi, S., Cole, G.D., Francis, D.P.:
  Automatic detection of end-diastolic and end-systolic frames in 2d
  echocardiography. Echocardiography  \textbf{34}(7),  956--967 (2017)

\end{thebibliography}


\section{Supplementary Material}

\subsection{Dataset properties}

The ACDC dataset was collected at two sites and covers adults with normal cardiac anatomy and four cardiac pathologies: systolic heart failure with infarction, dilated cardiomyopathy, hypertrophic cardiomyopathy and abnormal right ventricular volume. Each pathology is represented by 20 patients with pre-defined ES and ED phase. The CMR volumes have an average resolution of $220.12\pm{34.04} \times 247.14\pm{39.44} \times 9.51\pm{2.40}$ and a spacing of $1.51\pm{0.19} \times 1.51\pm{0.19} \times 9.34\pm{1.67}$ mm$^3$ (X/Y/Z). The resolutions and spacings (same ordering) are within in the following ranges: $[154,428]$, $[154,512]$, $[6,18]$, $[0.70,1.92]$,$[0.70,1.92]$,$[5,10]$. 
Each 4D CMR volume has between 84 and 450 2D slices with a mean of $253\pm{72}$.

The second cohort 
includes patients over 8 years with a complex congenital heart defect called Tetralogy of Fallot (TOF).  We used short-axis 4D cine series from 278 patients. They have an average resolution of $244.49\pm{60.02} \times 252.37\pm{48.82} \times 14.08\pm{3.20}$ and a spacing of $1.37\pm{0.18} \times 1.37\pm{0.18} \times 8.03\pm{1.25}$ mm$^3$ (X/Y/Z). The resolution and spacing (same ordering) are within the following min, max ranges: $[126,512]$, $[156,512]$, $[8,28]$, $[0.66,2.08]$,$[0.66,2.08]$,$[6,16]$. 
Each 4D CMR volume has between 112 and 700 2D slices with a mean of $313\pm{106}$ slices.

\small
\begin{table}
\centering
\caption{First row: Mean$\pm{SD}$, second row: min and max occurrence of the ground truth cardiac key frame indexes per ACDC and TOF dataset. Especially the cardiac key frame ranges in the TOF dataset, which comes from 14 different sites, illustrates the possible variability/permutations in clinical data.}
\label{tab:label_occurence}
\begin{tabular*}{\textwidth}{@{\extracolsep{\fill} }lccccc}
Data&ED&MS&ES&PF&MD\\
\toprule
ACDC
&20.42$\pm{12.26}$
&4.24$\pm{1.42}$
&9.91$\pm{2.67}$
&15.08$\pm{3.36}$
&22.92$\pm{15.14}$\\
&$[1:35]$
&$[3:12]$
&$[6:23]$
&$[9:30]$
&$[13:34]$\\
\midrule
TOF
&15.54$\pm{10.09}$
&4.91$\pm{1.72}$
&9.39$\pm{2.11}$
&13.22$\pm{12.41}$
&18.44$\pm{3.34}$\\
&$[1:34]$
&$[2:18]$
&$[2:22]$
&$[3:25]$
&$[7:30]$

\end{tabular*}
\end{table}

\subsection{Visual Examples}
\begin{figure}[h!]
\centering
  \includegraphics[width=\linewidth]{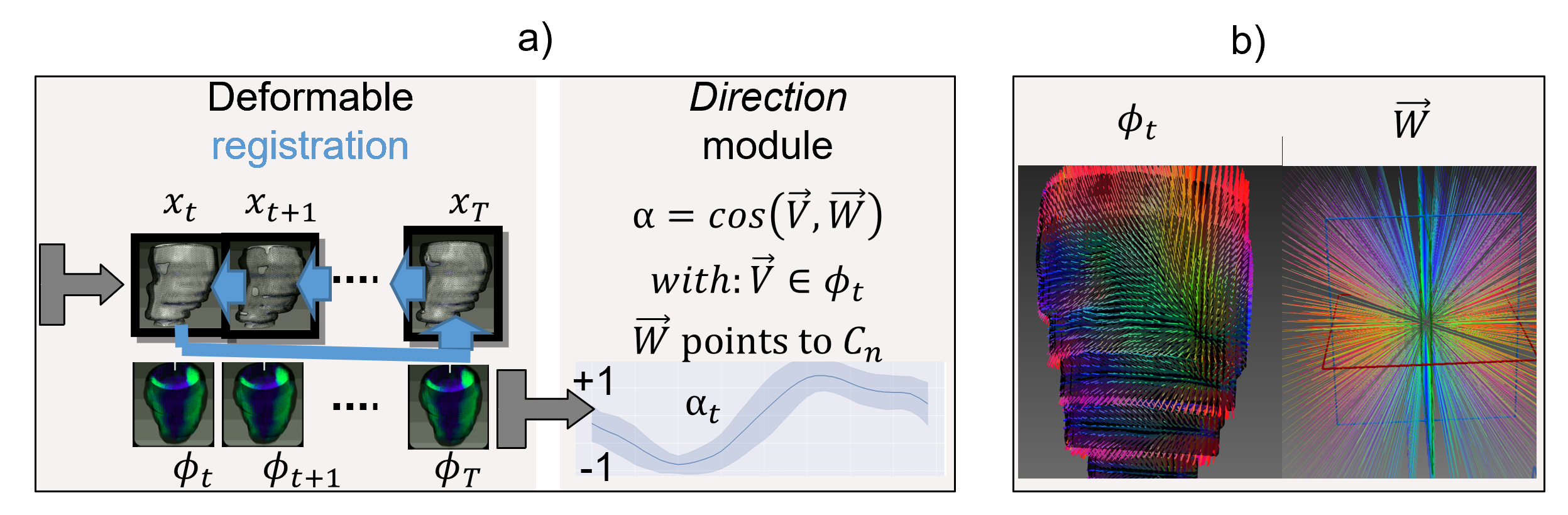}\\
  \caption{
  a) The registration module outputs $\phi$, which represents the deformation between neighbouring time steps $x_t$ and $x_{t+1}$. Following, the direction module which calculates the mean deformation angle/motion descriptor $\alpha_t$ between each vector $\vec{v} \in \phi$ and $\vec{w}$ pointing to $C_n$. 
  b) Example $\phi_t$ and the corresponding focus matrix with vectors $\vec{w}$ pointing to focus point $C_n$. Here, $\phi_t$ is masked by the left ventricle contours for visualisation purposes.}
  \includegraphics[width=1.\linewidth]{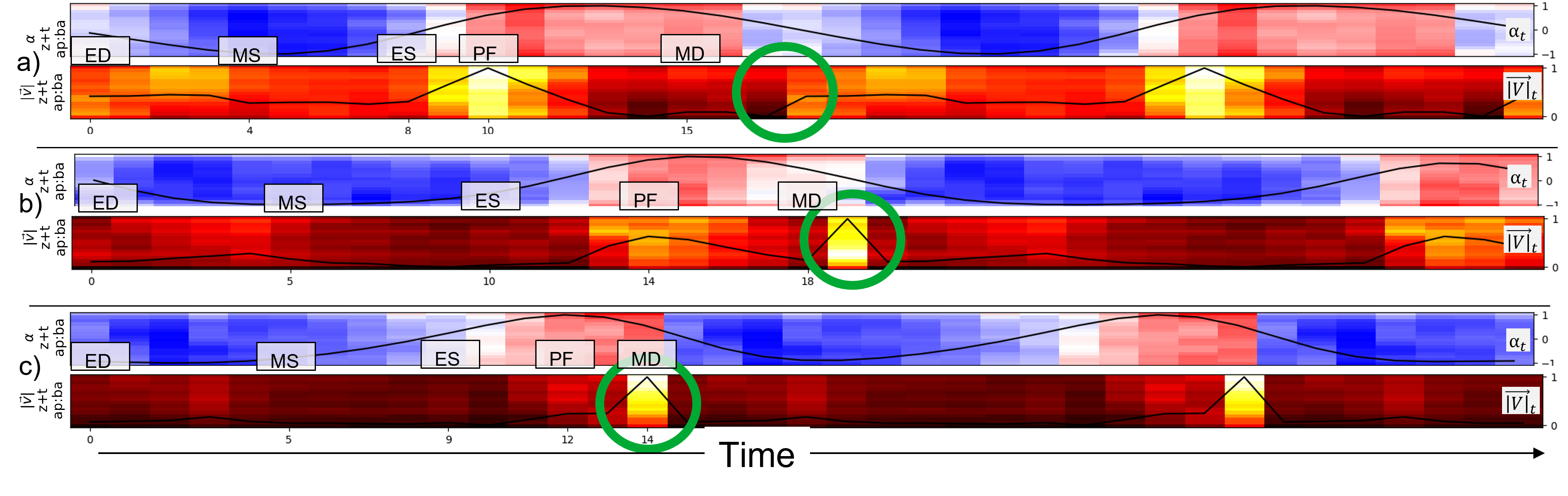}
  \caption{Motion descriptor ($\alpha_t$ and $|\vec{v}|_t$) derived from three patients, the green marker highlights the cycle-end; information is repeated until 40 time steps are filled (input format into network). A high peak reflects long deformation vectors in $|\vec{v}|_t$ towards the following time step. In the highlighted case it is $|\vec{v}|_T$, which is the deformation from the last volume $x_T$ to $x_1$. If $|\vec{v}|_t$ is an outlier with a high relative value compared to the other time steps it is a strong indicator for a cut-off CMR sequence. a) No cut-off. b) Moderate cut-off. c) Strong cut-off.}
 \includegraphics[width=1.\linewidth]{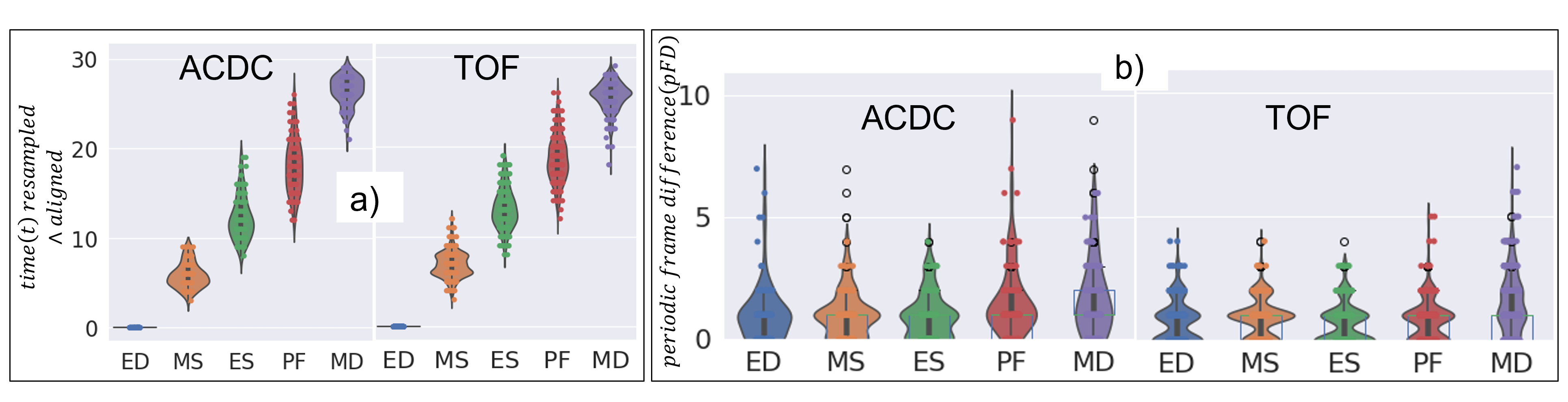}
 \caption{a) The distribution of the GT label show a much easier problem if we align and resize both datasets, which is not the case for clinical data (cf. Table 1). b) Average periodic frame difference (pFD) for each phase of the self-supervised method ($C_{mse}$) on the raw (no alignment/resizing) datasets.}
  \label{fig:dir}
\end{figure}
\end{document}